\pdfoutput=1 

\documentclass[11pt]{article}
\usepackage[]{emnlp2021}
\usepackage{nert}
\usepackage{times}
\usepackage{latexsym}
\usepackage{booktabs}
\usepackage{pifont}
\usepackage[section]{placeins}

\newcommand{\ftoken}[1]{\textcolor{orange}{\textbf{#1}}} 
\newcommand{\otoken}[1]{\textcolor{teal}{\textbf{#1}}} 

\newcommand{\cmark}{\ding{51}}%
\newcommand{\xmark}{\ding{55}}%
\newcommand{\lmark}{$\filledmedtriangleleft$}%
\newcommand{\rmark}{$\filledmedtriangleright$}%

\usepackage{microtype}

\newcommand{\quash}[1]{}  

\newcommand{\bx}{\mathbf{x}}
\newcommand{\bz}{\mathbf{z}}

\title{Putting Words in BERT's Mouth: 
\\ Navigating Contextualized Vector Spaces with Pseudowords}

\author{Taelin Karidi$^1$ \quad
Yichu Zhou$^2$ \quad
Nathan Schneider$^3$ \quad
Omri Abend$^1$ \quad
Vivek Srikumar$^2$ \\
$^1$Hebrew University of Jerusalem, \{\emldisplay{taelin.karidi@mail.huji.ac.il}{taelin.karidi}, \emldisplay{omri.abend@mail.huji.ac.il}{omri.abend}\}\texttt{@mail.huji.ac.il} \\
$^2$University of Utah, \{\emldisplay{flyaway@cs.utah.edu}{flyaway}, \emldisplay{svivek@cs.utah.edu}{svivek}\}\texttt{@cs.utah.edu} \\
$^3$Georgetown University, \eml{nathan.schneider@georgetown.edu} \\
}
\date{}

\begin{document}
\maketitle
\begin{abstract}

We present a method for exploring regions around individual points in a contextualized vector space (particularly, \textit{BERT space}), as a way to investigate how these regions correspond to word senses. 

By inducing a contextualized ``pseudoword'' as a stand-in for a static embedding in the input layer, 
and then performing masked prediction of a word in the sentence,
we are able to investigate the geometry of the BERT-space in a controlled manner around individual instances. 
Using our method on a set of carefully constructed sentences targeting ambiguous English words, we find substantial regularity in the contextualized space, with regions that correspond to distinct word senses; but between these regions there are occasionally ``sense voids''---regions that do not correspond to any intelligible sense.\footnote{Our code and dataset are available at \url{https://github.com/tai314159/PWIBM-Putting-Words-in-Bert-s-Mouth}}
\end{abstract}

\section{Introduction}

Vector spaces defined over static word vectors are somewhat interpretable, as the points are limited to the vocabulary.
Contextualized representations (CRs), by contrast, are mysterious because of the unbounded number of distinct contextualized embeddings, and no obvious way to discover the word and context that would correspond to an arbitrary point in the space.
Attempts have been made to characterize the information captured in contextualized representations \cite{rogers-etal-2020-primer,tenney2019b,liu-etal-2019-linguistic}, but some of the techniques used (e.g., probing classifiers) have been subject to criticism for their indirectness. 

We propose a new technique called \textbf{Masked Pseudoword Probing} (MaPP) that allows controlled exploration of the space of a contextualized masked LMs \citep[specifically, English BERT;][]{devlin-etal-2019-bert}.
MaPP takes advantage of the static embedding at the first layer of BERT and ``hallucinates'' new embeddings into this space to correspond to tokens' contextualized representations.
By extending BERT's vocabulary with these \emph{pseudowords}, we can use them as inputs for masked prediction of words in the sentence.
The words predicted in the masked slot serve as evidence of the meaning of the pseudoword in question---for example, it may encapsulate a specific sense consistent with the original context. 
We can also transplant a pseudoword into new contexts to see if it generalizes as per our intuitions about word meanings.

\begin{figure}
\centering
\includegraphics[width=\columnwidth]{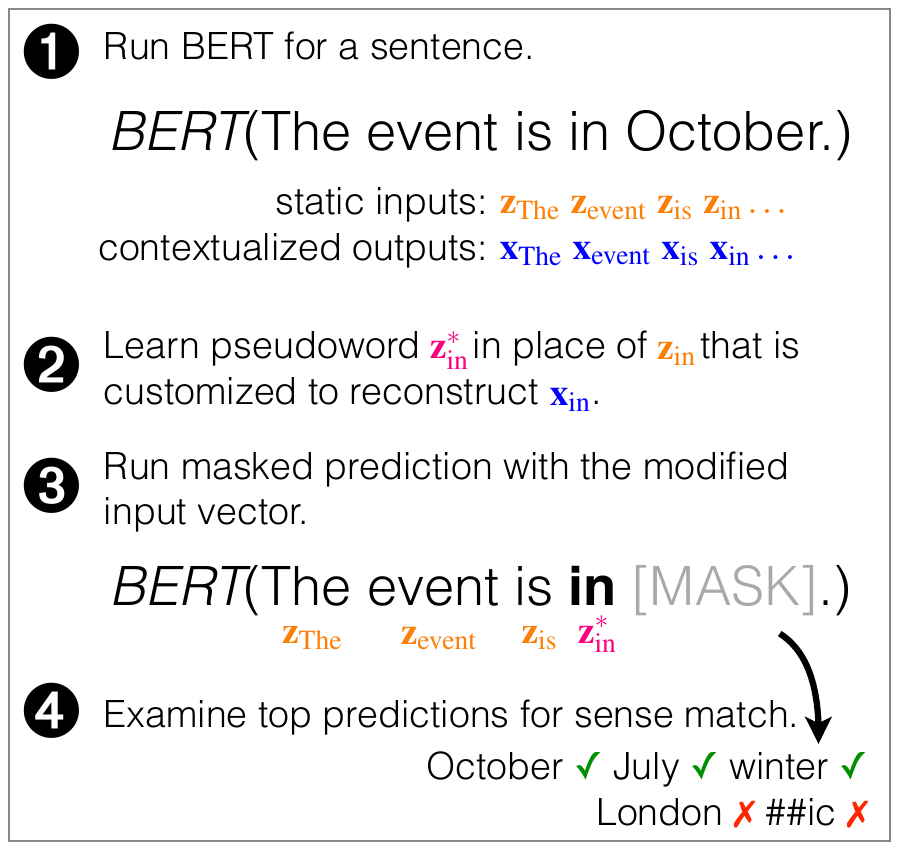}
\caption{Illustration of the MaPP method as used in the specialization experiments (\cref{sec:specialization}). In other experiments, the pseudoword is perturbed prior to step~3.}\label{fig:approach}
\end{figure}

We focus on the contextualized meanings of ambiguous verbs and prepositions, which serve as ideal test cases, as they may often be disambiguated by their objects. For example, if a pseudoword interpreted as ``in'' induces a distribution over its argument slot that gives most of the probability mass to locations such as ``London'' or ``Paris'', we may conclude that the pseudoword has a locative sense (\cref{tab:main_examples}). 
We first ask: How well do pseudowords inferred from contextualized representations in BERT accord with linguistic expectations about word senses? 
We investigate this by deploying MaPP for 
sentences whose masked slot reveals the sense of the ambiguous word.
Second, we ask: How semantically smooth is the BERT-space such that arbitrary points near a pseudoword will behave in semantically similar ways?
We study this by navigating the space around a pseudoword (or between two pseudowords), and examining the vector's behavior via MaPP. 
This method allows for investigating the geometry of a contextual representation by traveling the BERT-space in a continuous way and exploring different regions, which we show (see \cref{section_experiments}) to correspond to distinct concepts. 

Our experiments indicate a substantial regularity in the BERT-space. We see regions in the space that correspond to distinct senses. These regions can be recovered using our technique; for example, by sampling points around a pseudoword and looking at the points in the BERT-space which decode to it. Moreover, we see that between sense-regions there are often ``voids'' in the space that do not correspond to any intelligible sense.


\section{Analyzing Contextual Representations} \label{sec:analyzingCR}

\paragraph{Probing representations.}

Deciphering the information encoded in contextualized representations like BERT is widely investigated in recent NLP research. 
\emph{Probing} methods use CRs as inputs to probing classifiers to see how well the CRs may serve as features in predicting specific properties.
The intuition is that if the CR can be used to predict a specific property, then  knowledge about it is encoded in the representation. 
Recent classifier-based probes have focused on various linguistic properties such as morphology, parts of speech, sentence length, and syntactic and semantic relations \citep[][\emph{inter alia}]{liu-etal-2019-linguistic,conneau-etal-2018-cram,belinkov-etal-2017-neural,Adi17-morphology-probing}. 
Closely related to ours is work that studied 
the extent to which the lexical semantic classes of nouns are disambiguated by CRs
\citep{zhao-etal-2020-quantifying}, showing that BERT fares well in this respect.

Beyond classifier-based probes, other approaches have also been explored, such as information theoretic probing~\citep{pimentel-etal-2020-information,voita-titov-2020-information}, and structural probing~\citep{hewitt-manning-2019-structural}, which evaluates whether syntax trees are embedded in a linear transformation of a CR's word representation space. 

An alternative approach to probing learned representations is directly analyzing the attention weights and activation patterns~\citep{brunner-etal-2020-corpus, abnar-zuidema-2020-quantifying}. Criticism against some instances of this approach is found in \citet{Jain2019}, who claimed that attention weights are less transparent than is often stipulated.

\paragraph{The shortcomings of probing.}

Some recent work has taken a more critical view regarding probing techniques \citep{belinkov-probing2021}. \citet{amnesicprob} argue that while probing methods might show that certain linguistic properties exist in a representation, they do not reveal how and if this information is being used by the probing model. 
This could be due to the disconnect between the representation itself and the probing model. 

Relying on classifiers to interpret representations might be problematic; they add additional confounds to the interpretability of the results, and different representations may need different classifiers~\cite{wu-etal-2020-perturbed,zhou-directprobe}.

Another critique concerns the difference between correlation and causation~\cite{Feder20-causal-lm}: classifier-based probes may rely on shallow correlations in the training set, thus reflecting data artifacts that are irrelevant to the studied distinction. 

\paragraph{Word Sense Disambiguation.}

Word Sense Disambiguation (WSD) aims at making explicit the semantics of a word in context, typically by identifying the most suitable meaning from a predefined sense inventory \citep{Bevilacqua21}. Disambiguation can also be defined indirectly, through minimal pairs that contrast two senses of a word \citep{trott-bergen-2021-raw} or through another word in the text that determines the semantic class of the word in question \citep{jiang-riloff-2021}.

Our work bears on this line of work as well:

we are using MaPP to test whether the masked prediction indicates that the pseudoword encodes the expected sense. However, we are using carefully controlled sentences so it remains to be seen whether pseudowords can be induced to capture word senses “in the wild”.


\paragraph{The geometry of BERT.} Understanding the geometry of the BERT-space is not  easy. Some attempts in this direction have been made~\citep{coenen-reif-yuan19,ethayarajh-19,michael-etal-2020-asking, mickus-etal-2020-mean, xypolopoulos-etal-2021-unsupervised, gari-20}, but a more thorough investigation is lacking. As opposed to \textit{predictive methods} such as probing, \textit{descriptive methods} that rely on geometric features of the space analyze the information in CRs directly. 

This paper takes a different approach that views BERT as a function that is defined over a continuous space.
Our proposed methodology thus allows for a more direct inspection of “gaps” between embedded tokens, that does not require an auxiliary probe and probing dataset, and instead investigates the model’s behavior on arbitrary points in the input space. The paper focuses then on the interpretation of individual points, as opposed to other related work on the geometry of BERT~\cite{coenen-reif-yuan19,ethayarajh-19,cai2021isotropy, Hernandez2021TheLL}, which mostly considers higher-level properties of the BERT space.

A na\"{i}ve geometric approach to investigate the information in BERT could be to look at neighborhoods of contextualized embeddings.
A vector in this space represents some word within a sentence; it lies in $\mathbb{R}^{d}$, with $d = 768$.
However, it is unclear how such neighborhoods should be defined.

Of course, it is possible to define a discrete neighborhood comprising of contextualized embeddings close to the vector; these may represent the same sense of the same word. 
Still, in terms of the geometry of the space, how should we interpret a \emph{continuous} neighborhood in the output space? While we could force a non-discrete outlook by generating vectors artificially---e.g., by generating points that are epsilon away from a given point---these artificial contextualized vectors are disembodied, with no obvious linguistic basis. It is therefore unclear how to interpret these artificial vectors (or the linguistic properties of tokens they might encode).

\section{Traversing CR Spaces: A New Probing Methodology}

Our motivating question is: Are word senses encoded in BERT's representations---and if so, how? As a test case, we look at highly ambiguous words, as they potentially offer complex geometric configurations of senses in the BERT space.

\begin{table*}[!t]
\centering
\footnotesize
\renewcommand{\tabcolsep}{0.15cm}
\begin{tabular}{cllp{10em}p{7.5em}}
\toprule
\textbf{Focus Word} & 
\multicolumn{1}{c}{\textbf{Sentence A}} & 
\multicolumn{1}{c}{\textbf{Sentence B}} & 
\multicolumn{1}{c}{\textbf{Sense A}} & 
\multicolumn{1}{c}{\textbf{Sense B}}\\
\midrule
in & The event is \ftoken{in} \otoken{October}.  & The event is \ftoken{in} \otoken{London}. & temporal & locative \\ 
for & The book is \ftoken{for} \otoken{Lisa}.  & The book is \ftoken{for} \otoken{reading}. & person & purpose \\
with & I ate salad \ftoken{with} \otoken{enjoyment}.  & I ate salad \ftoken{with} a \otoken{knife}. & feeling & instrument \\ 
 
about & The clip is \ftoken{about} a \otoken{horse}.  & The clip is \ftoken{about} a \otoken{minute}. & topic & duration \\[5pt]

started & I \ftoken{started} the \otoken{car}.  & I \ftoken{started} the \otoken{book}. & device & information source \\
had & I \ftoken{had} a \otoken{party}.  & I \ftoken{had} a \otoken{fever}. & social event & medical condition \\ 
had & I \ftoken{had} \otoken{slept}.  & I \ftoken{had} \otoken{pizza}. & auxiliary/past participle & food \\ 
 electronic devices \\ 
\bottomrule
\end{tabular}
\caption{Example minimal pairs from our dataset. Each pair differs in the ambiguous word's argument (and determiner if needed), such that the  ambiguous word holds a different sense.}
\label{table min pairs}
\label{tab:main_examples}
\end{table*}

\subsection{Masked Pseudoword Probing} \label{sec:traversing_bert}

We propose a novel probing technique, \emph{MaPP (Masked Pseudoword Probing)},  which ``hallucinates'' vectors 
to reconstruct a token's contextualized representation. 
MaPP allows us to ``navigate'' the BERT-space by looking at neighborhoods of certain word vectors 
in what we term the \textit{input space}, an extension of the discrete space of BERT's static (decontextualized) word embeddings.
By inducing and manipulating new vectors in the input space (henceforth, \textbf{pseudowords}), we can observe the effects on BERT's behavior via masked prediction.
Mathematically, pseudowords are inverse images under the BERT function, continued from the finite space of word embeddings that BERT generally receives, which in the standard implementation contains 30k points in $\mathbb{R}^{d}$, 
one per entry in BERT's vocabulary. \footnote{The term ``pseudoword'' is also used in psycholinguistics, but refers to a different concept \citep{gale92,schutze-1998-automatic,shoemark-etal-2019-room}.}

We are interested in the contextualized representation of an ambiguous word, which we call the \textbf{focus token} $t$ in a sentence $s$.

Let $\bz_t$ be the static embedding of $t$ (i.e., the input embedding BERT receives for $t$), and let $\bx_t$ be its contextualized representation in $s$. Let $d$ be the dimension of the input embeddings (without the positional embedding).

To apply our method, we stipulate that there is a specific token called the \textbf{cue token}, in the $j^{\text{th}}$ position in $s$, 
that
disambiguates $t$. Under this assumption, MaPP can discover the sense of a vector $\bx$ in the vicinity of $\bx_t$, by masking the cue token, and decoding the distribution of its fillers. These fillers serve as a proxy for $\bx$'s sense.

MaPP operates in two steps, described next.

\paragraph{1) Pseudoword Representation.}

First, we find an embedding $\bz^*$ that best reconstructs $\bx_t$ when input to BERT in place of $t$. 
Formally: 
\begin{equation} \label{loss}
\bz^{*} = \argmin_{\bz \in \mathbb{R}^{d}} {|| BERT(\bz) -  \bx_t ||}^2 
\end{equation}
where $BERT(\bz)$ is a forward pass of the model with the vector $\bz$ replacing $\bz_t$. The solution  $\bz^*$ is a pseudoword.
The original embedding $\bz_t$ of $t$ is a solution to \cref{loss}. But BERT is not invertible, and there is no reason for $\bz^*$ and $\bz_t$ to be close. Indeed, 
our experiments show that $\bz^*$ is different from $t$'s input embedding,\footnote{We looked at the distribution of the distances (Euclidean and cosine) between all of the pseudowords and their corresponding static embeddings in the input space.} and suggest that $\bz^*$ is a ``disambiguated'' counterpart of the focus token.

We can approximate $\bz^*$ using standard optimization techniques. When solving for $\bz^*$, we hold BERT's parameters fixed, and seek to identify the input embedding $\bz$. In standard BERT training, by contrast, the input is known, and we solve for BERT's parameters.

\paragraph{2) Pseudoword-Guided Prediction.}

After computing $\bz^*$, we define a new sentence $s'$, identical to $s$ except for the $j^{\text{th}}$ position (the disambiguating position), where we place  a mask. For example:

\ex.\a.\label{coer:ss_loc} $s$: The event is \ftoken{in} \otoken{London}.
    \b.\label{coer:ss_temp} $s'$: The event is \ftoken{in} [MASK].
    \z.

The focus token $t$ is the ambiguous ``in''. The cue tokens ``London'' and ``September'' would indicate locative and temporal senses of ``in'' respectively.

Next, we replace the input embedding of $t$ with $\bz^*$ or another input space vector $\bz$ in its vicinity, and predict the distribution of the slot fillers in the masked position.\footnote{We also verify that the $\bz$ indeed decodes to $t$, i.e., when $\bz^*$ replaces the focus token's embedding, the resulting probability is concentrated on ``in''.}

\subsection{Pseudowords as Input Vectors.} 
\label{hallucinated_as_input} 

Let us denote the standard input space for the token at $t$'s position (i.e., input embeddings of BERT's vocabulary) with $I_{\textnormal{static}} \subset \mathbb{R}^{d}$ (where $|I_{\textnormal{static}}| = 30\text{k}$).   
By extending BERT's inputs to pseudowords, we are performing what is known in mathematical analysis as a \textit{continuation} of BERT's function from the discrete input space $I_{\textnormal{static}} \subset \mathbb{R}^{d}$ to continuous regions in $\mathbb{R}^{d}$. 
This approach allows us to gain insight as to the semantics encoded by different regions of the BERT space. 
Construing BERT as a continuous function also allows us to invert it, and obtain a point in the inverse image $z^*$ of BERT by solving an optimization problem.
We note that viewing the BERT space as a continuous space, e.g., for purposes of mapping between it and other continuous spaces, is an increasingly common practice \cite{schuster-etal-2019-cross, gauthier-levy-2019-linking}; see further discussion in \cref{app:further_discussion}. %
In our experiments (\cref{section_experiments}), the pseudowords will help us explore the geometry of the BERT-space, by traveling across it in a ``continuous'' way---something that is not possible to do with the BERT vectors as discussed in \cref{sec:analyzingCR}. 
For example, we can study how perturbations of $\bz^*$ (the pseudowords) affect the  prediction of the cue word.

\section{Experimental Research Questions $\&$ The MaPP Dataset} \label{sec:hypothesis}

The main hypothesis we study in this paper is:

\begin{quote}
There are regular ``nicely defined'' regions in the BERT-space around words that correspond to distinct senses. 

\end{quote}
Such regions may be variously interpreted: Around a point with a particular sense, there is a ball which contains mostly points corresponding to that sense. Or, for example, points that correspond to the same sense will lie on a high-dimensional manifold. 
Cases that we consider as not ``nicely defined'', are, for example, points corresponding to different senses that are scattered in the space in an inseparable way (or at least inseparable by simple functions).
We would like to be able to map the semantic concept of sense to the geometric properties of the BERT-space. 
However, since little is known about the space, fully characterizing its geometry is out of the scope for this work.

We present MaPP to study this hypothesis, and in doing so introduce the concept of pseudowords. This concept opens additional research questions.

\paragraph{Specialization.} Let $\bz^* \in \mathbb{R}^d$ be a pseudoword obtained by solving \cref{loss} for a sentence $s$ with a focus token $t$ and cue token at position $j$, holding a sense $\eta$. 
Does $\bz^*$ yield a sense distribution (determined by its slot fillers in the $j^{\text{th}}$ position) that concentrates on $\eta$? 
That is, does a pseudoword decode to a specific sense of the focus token?

\paragraph{Generalization.} Is it possible to transplant a pseudoword into a sentence where the context around the focus token is different, and still obtain coherent results? 
For example, in the sentences: (a) ``The pan is \ftoken{for} \otoken{cooking}.''\ and (b) ``The fork is \ftoken{for} \otoken{eating}?'', the focus token ``for'' is the same, and in both cases has a \textsc{purpose} meaning. The context, however, is different. If we transplant the pseudoword for ``for'' induced with sentence (a), to the position of ``for'' in a masked version of sentence (b) (i.e, ``The pan is \ftoken{for} [MASK]?''), will we get a coherent prediction with the same sense? Or is the pseudoword obtained from one sentence limited to a specific context? 

If pseudowords obtained for one sentence do not generalize to another, we propose to induce a ``generalized pseudoword'' trained over multiple examples with the same sense.

\subsection{The MaPP Dataset}\label{sec:dataset}

To answer the questions listed above, we manually compiled the MaPP Dataset, a controlled dataset with short sentences, designed to avoid confounds that may introduce difficulties in interpreting the results.

Each sentence contains an ambiguous word  that is fully disambiguated by a specific slot in the sentence. E.g., in the sentence ``The book is \ftoken{for} \otoken{reading}'', the ambiguous word ``for'' has a \textsc{purpose} sense, strongly signaled by ``reading''. 
All sentences were reviewed by a linguist to maximize naturalness and minimize ambiguity.

The dataset consists of 3 portions, each used in different experiments. We describe each portion adjacent to the relevant experiment.

\paragraph{Relational words as a test case.}
We chose to focus our analysis on the ambiguity of relational words in English, specifically prepositions and verbs. Relational words present an interesting test case: many are highly ambiguous and encode basic semantic distinctions, such as space, time and manner \cite{schneider-etal-2018-comprehensive}.
We do not attempt to cover all possible senses of the selected words; 
instead, we have constructed our dataset to illustrate just a few clear contrasts (see further discussion in \cref{app:further_discussion}).


\section{Experiments} \label{section_experiments}

We use MaPP to empirically evaluate the hypotheses listed above. 

We conduct four types of experiments. 
First, we test specialization, the extent to which the induced pseudoword $\bz^*$ can be viewed as a sense-disambiguated version of the focus token's input embedding. 
Second, we venture into the immediate regions around $\bz^*$ by perturbing it, and examining how this affects the resulting senses. We thereby gain insight as to the regularity of the region around $\bz^*$ with respect to the focus token's sense.
Third, we examine the sense regularity of the BERT CRs by examining the senses encountered when traversing the line between two pseudowords corresponding to different senses (e.g., the locative and temporal senses of ``in''). 
Finally, we test generalization, namely, the extent to which $\bz^*$ can serve as a sense-disambiguated embedding of the focus word \emph{in different contexts}.

\subsection{Implementation Details}\label{sec:implementation}

Throughout, we use the BERT-base-cased model via the implementation in HuggingFace~\cite{wolf-etal-2020-transformers} and Pytorch~\cite{NEURIPS2019_9015}.

To solve for $\bz^*$ (\cref{loss}), we add a new token to the vocabulary (\texttt{\#TOKEN\#}), which corresponds to the \textit{focus token}. When backpropagating the gradients, we ensure that the gradients of all parameters of BERT are zero, except the token embedding of \texttt{\#TOKEN\#}. In this way, we preserve the original BERT model while enabling us to solve for $\bz^{*}$. We use $5$ random initializations and select the $z^*$ with the lowest loss.
We use standard gradient-based optimization for this process. We are solving for the input to BERT rather than model parameters, so we backpropagate through the network, holding BERT's parameters fixed, and take gradients with respect to $z$.

 
\subsection{Specialization Experiments}\label{sec:specialization}


\begin{table}[t]
\centering
\small
\resizebox{\columnwidth}{!}{
\begin{tabular}{@{}>{\raggedright}p{6.25em}@{~~~}l@{~}p{22em}@{}}
\toprule
\multicolumn{1}{c}{\textbf{Query}} & & \multicolumn{1}{c}{\textbf{Top 5 predictions}}  \\
\midrule

\multirow{2}{6.25em}{The dinner is \ftoken{on} \otoken{Monday}.}
& $\bz$   & 
fire  \xmark{} 
offer  \xmark{} 
sale  \xmark{} 
\mbox{Friday \cmark{}}
hold  \xmark{} \\
& $\bz^*$ & 
Sunday \cmark{}
Saturday  \cmark{}
Thursday  \cmark{} 
Tuesday  \cmark{} 
Friday  \cmark{} 
\\
\midrule

\multirow{2}{6.25em}{The clip is \ftoken{about} a \otoken{queen}.}
& $\bz$ & 
minute  \xmark{}
year  \xmark{}
second  \xmark{}
day  \xmark{}
week  \xmark{}  \\
& $\bz^*$ & 
woman  \cmark{} 
girl  \cmark{} 
man  \cmark{} 
child  \cmark{} 
boy \cmark{} 
\\
\bottomrule
\end{tabular}
}
\caption{Specialization examples where the pseudoword $\bz^*$ learned from the query sentence corresponds to a different sense from BERT's static word embedding $\bz$, as evidenced by the top~5 predictions when the cue token (\otoken{Monday}, \otoken{queen}) is masked out. }
\label{tab:spec_examples}
\end{table}

We test whether we can control the sense of BERT's predicted tokens using a \textit{pseudoword} $\bz^*$. If this is indeed the case, it supports our view of $\bz^*$ as a disambiguated version of its corresponding static token embedding. 
In this experiment, we designate highly ambiguous words---specifically, verbs like ``have'' and prepositions like ``in''---as the focus tokens and apply the process described in~\cref{sec:traversing_bert} on these ambiguous words.  

\paragraph{Data.}
We use the \textbf{Basic Portion} of the MaPP Dataset for this experiment. It contains 94~sentences with $8$ ambiguous words: \w{had}, \w{started}, \w{run}, \w{in}, \w{for}, \w{with}, \w{about}, and \w{on}. Two to four senses of each word type are represented with 5~sentences each (except for rare cases where we could not find five coherent sentences for a certain template).

\paragraph{Evaluation.} For each sentence 
we perform masked prediction and manually categorize which of the top 5 predicted words are consistent with the original sense. 
From these judgments we compute the accuracy---the proportion of sense-congruent predictions\footnote{We also conducted a random baseline experiment, with randomly sampled vectors from $\mathbb{R}^d$ instead of the pseudoword. The accuracy was negligible.}. 
In total, we evaluate 470 predictions.

\paragraph{Results.}
\Cref{table_specialization} shows the performance of MaPP versus a Vanilla BERT baseline (using the static embeddings rather than pseudowords for masked prediction). 
We see that in most cases, by applying MaPP, we shift the prediction of the model to the desired sense, which establishes the validity of our technique.  
Further, \cref{table_specialization_mask} shows that typically, after applying MaPP, the model's top prediction is not the word that was masked in the original sentence---i.e., the pseudoword is not simply memorizing a specific cue word.
\Cref{tab:spec_examples} illustrates two examples exhibiting a clear shift to the desired sense. 
Not every pseudoword behaves as expected, though, 
as is discussed in subsequent experiments.

\begin{table}[t] 
\centering
\small
\resizebox{\columnwidth}{!}{
\begin{tabular}{lrrrrrrr}
\toprule
\textsc{Sense Match} & \multicolumn{2}{c}{All} & \multicolumn{2}{c}{Verbs} & \multicolumn{2}{c}{Prepositions} \\
 \cmidrule(lr){2-3}\cmidrule(lr){4-5}\cmidrule(lr){6-7}
 & \textbf{@1} & \textbf{@5} & \textbf{@1} & \textbf{@5} & \textbf{@1} & \textbf{@5}\\
$N$ (total \# of predictions) & 94 & 470 & 43 & 215 & 51 & 255 \\
\midrule
Vanilla BERT ($\bz$) & 39.8 & 36.0 & 36.4 & 27.9 & 41.0 & 42.7 \\
MaPP ($\bz^*$) & 77.6 & 65.1 & 83.7 & 67.0 & 72.5 & 63.5 \\
 
\bottomrule
\end{tabular}
}
\caption{Specialization results: accuracy at producing a completion consistent with the sense from the original context (out of top-1 and top-5 predictions). Subscores are provided for verb \&~preposition focus words.}
\label{table_specialization} 
\end{table}

\begin{table}[t] 
\centering
\small
\resizebox{\columnwidth}{!}{
\begin{tabular}{@{}lrrrrrrrrr@{}}
\toprule
\textsc{Word Match} & \multicolumn{3}{c}{All} & \multicolumn{3}{c}{Verbs} & \multicolumn{3}{c@{}}{Prepositions} \\
 \cmidrule(lr){2-4}\cmidrule(lr){5-7}\cmidrule(l){8-10}
 & \textbf{@1} & \textbf{@5} & \textbf{@20} & \textbf{@1} & \textbf{@5} & \textbf{@20} & \textbf{@1} & \textbf{@5} & \textbf{@20} \\
$N$ & 94 & 470 & 1880 & 43 & 215 & 860 & 51 & 255 & 1020 \\
\midrule
Vanilla BERT & 5.3 & 17.0 & 35.1 & 2.3 & 11.6 & 20.9 & 7.8 & 21.6 & 47.1  \\
MaPP & 24.5 & 51.1 & 74.5 & 25.6 & 55.8 & 79.1 & 23.5 & 47.1 & 70.6  \\
  
\bottomrule
\end{tabular}
}
\caption{Specialization results: rate of predicting the word that was masked in the original sentence (recall in top $k$ predictions).}
\label{table_specialization_mask} 
\end{table}

\subsection {$\epsilon$-Perturbation}\label{ssec:eps_pert}

\begin{figure}[t]
    \centering
    \includegraphics[width=0.25\textwidth]{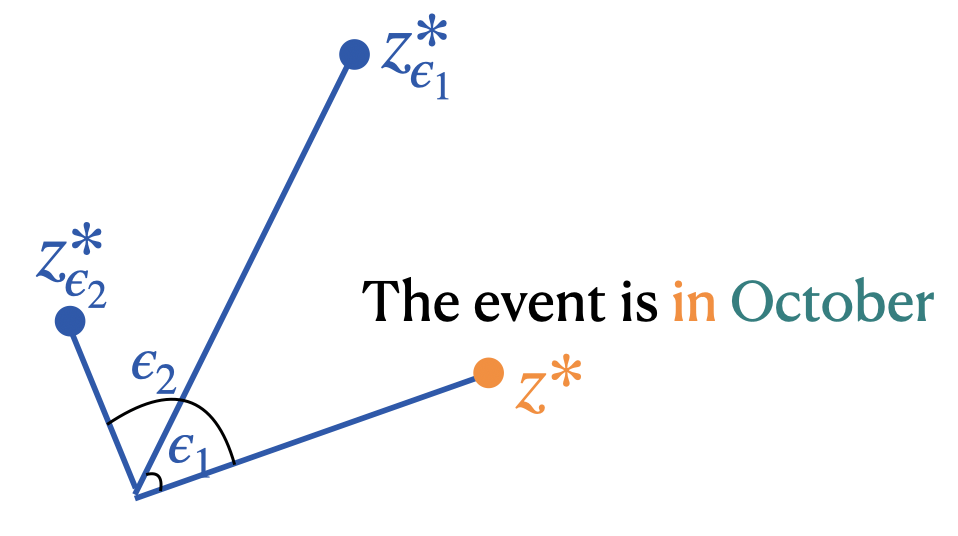}
    \caption{Perturbation of $\bz^{*}$ by different $\epsilon$ values.}
    \label{fig:perturbation}
\end{figure}

Our goal in this experiment is to ``travel'' in the BERT-space. Since it is not clear how to interpret a direct perturbation of the contextualized vectors, we do this via the input space. We compute a pseudoword and perturb it to obtain new points in an $\epsilon$-ball around it, as schematized in \cref{fig:perturbation}.

Given a pseudoword $\bz^{*}$ for a particular token in context, normalize it to a unit vector. Choose 10~random directions $w$ by sampling uniformly from the unit sphere. For each direction $w$ and perturbation distance $\epsilon$, find a vector $w'$ that is $\epsilon$ away (in cosine distance) from $\bz^{*}$ in the direction $w$ (i.e., $w'$ is on the intersection between the plane spanned by $\bz^{*}$ and $w$, and the unit sphere). We do this for several values of $\epsilon$ (see below). Each perturbed $\bz^*$ is fed back into the model and used for masked prediction.

\begin{figure}[t] 
\centering

\includegraphics[clip,trim=0 0 0 70,width=7cm]{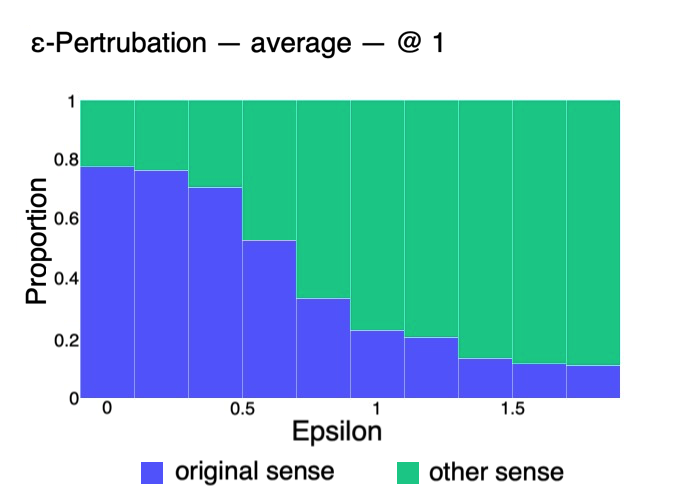}
\caption{$\epsilon$-perturbation: average accuracy at producing a completion consistent with the sense from the original context, for each $\epsilon$ (average over $10$ directions per $\epsilon$). 
On average, as $\epsilon$ increases, the rate of matching the original sense decreases.
\vspace{-.5cm}}
\label{plot_bar_epsilon_top1}
\end{figure}

\paragraph{Data.} In this experiment we use the \textbf{Basic Portion} of the MaPP Dataset, as described in~\cref{sec:specialization}. 

\paragraph{Evaluation.} For $\epsilon \in \{0,0.2,...,1.8 \}$ and each direction, we examine at the model's top 5 predictions for the masked token to determine which are consistent with the original sense. We measure the average accuracy over the $10$ directions for each $\epsilon$.


\begin{table*}[!t]
\centering
\small
\resizebox{\textwidth}{!}{
 \begin{tabular}{|>{\raggedright\centering}p{9.75em}|c|>{\raggedright\centering}p{10em}|c|c|c|c|>{\raggedright\centering}p{9.5em}@{}r|@{}}
	\hline

\multirow{2}{*}{\textbf{Mask}} & \multirow{2}{*}{\textbf{Vanilla BERT}} & \multirow{2}{*}{\textbf{Query 1}}  &
 \multicolumn{4}{c|}{\textbf{Interpolated MaPP}} & \multirow{2}{*}{\textbf{Query 2}} & \\ \cline{4-7}
	
	  &  &  & \textbf{$\alpha = 0$} & \textbf{$\alpha = 0.4$} & \textbf{$\alpha = 0.8$} & \textbf{$\alpha = 1$} & &
 \\ \hline

The event is in [MASK]. & \begin{tabular}{@{}p{5em}@{}c@{}} 
progress & \xmark \\
June & \rmark \\
July & \rmark \\
April & \rmark \\
September & \rmark \\
\end{tabular} & The event is \ftoken{in} \otoken{London}.
  & \begin{tabular}{@{}p{5em}@{}c@{}} 
\textbf{London} & \lmark \\
Dublin & \lmark \\
Edinburgh & \lmark \\
Paris & \lmark \\
Sydney & \lmark \\
\end{tabular} & \begin{tabular}{@{}p{5em}@{}c@{}} 
Toronto & \lmark \\
\textbf{London} & \lmark \\
June & \lmark \\
Dublin & \lmark \\
Melbourne & \lmark \\
\end{tabular} & \begin{tabular}{@{}p{5em}@{}c@{}} 
June & \rmark \\
July & \rmark \\
March & \rmark \\
September & \rmark \\
April & \rmark \\
\end{tabular} & \begin{tabular}{@{}p{5em}@{}c@{}} 
July & \rmark \\
September & \rmark \\
June & \rmark \\
March & \rmark \\
\textbf{August} & \rmark \\
\end{tabular}  & The event is \ftoken{in} \otoken{August}. & \\

\hline
The book is for [MASK]. & \begin{tabular}{@{}p{5em}@{}c@{}} 
children & \lmark \\
women & \lmark \\
adults & \lmark \\
sale & \rmark \\
boys & \lmark \\
\end{tabular} & The book is \ftoken{for} \otoken{him}.
  & \begin{tabular}{@{}p{5em}@{}c@{}} 
me & \lmark \\
her & \lmark \\
\textbf{him} & \lmark \\
you & \lmark \\
us & \lmark \\
\end{tabular} & \begin{tabular}{@{}p{5em}@{}c@{}} 
children & \lmark \\
women & \lmark \\
you & \lmark \\
sale & \rmark \\
free & \xmark \\
\end{tabular} & \begin{tabular}{@{}p{5em}@{}c@{}} 
free & \xmark \\
sale & \rmark \\
download & \rmark \\
reading & \rmark \\
children & \lmark \\
\end{tabular} & \begin{tabular}{@{}p{5em}@{}c@{}} 
free & \xmark \\
download & \rmark \\
sale & \rmark \\
reading & \rmark \\
purchase & \rmark \\
\end{tabular}  & The book is \ftoken{for} \otoken{viewing}. & \\

\hline
I started the [MASK]. & \begin{tabular}{@{}p{5em}@{}c@{}} 
engine & \lmark \\
\textbf{car} & \lmark \\
motor & \lmark \\
truck & \lmark \\
ignition & \lmark \\
\end{tabular} & I \ftoken{started} the \otoken{car}.
  & \begin{tabular}{@{}p{5em}@{}c@{}} 
engine & \lmark \\
\textbf{car} & \lmark \\
ignition & \lmark \\
truck & \lmark \\
motor & \lmark \\
\end{tabular} & \begin{tabular}{@{}p{5em}@{}c@{}} 
engine & \lmark \\
\textbf{car} & \lmark \\
ignition & \lmark \\
truck & \lmark \\
motor & \lmark \\
\end{tabular} & \begin{tabular}{@{}p{5em}@{}c@{}} 
engine & \lmark \\
\textbf{car} & \lmark \\
\textbf{book} & \rmark \\
machine & \lmark \\
fire & \xmark \\
\end{tabular} & \begin{tabular}{@{}p{5em}@{}c@{}} 
\textbf{book} & \rmark \\
story & \rmark \\
game & \rmark \\
engine & \lmark \\
movie & \rmark \\
\end{tabular}  & I \ftoken{started} the \otoken{book}. & \\

\hline
I had [MASK]. & \begin{tabular}{@{}p{5em}@{}c@{}} 
to & \xmark \\
it & \xmark \\
nothing & \xmark \\
him & \xmark \\
her & \xmark \\

\end{tabular} & I \ftoken{had} \otoken{cake}.
  & \begin{tabular}{@{}p{5em}@{}c@{}} 
  
anxiety & \xmark \\
it & \xmark \\
worry & \xmark \\
energy & \xmark \\
power & \xmark \\

\end{tabular} & \begin{tabular}{@{}p{5em}@{}c@{}} 

it & \xmark \\
doubts & \xmark \\
to & \xmark \\
problems & \xmark \\
anxiety & \xmark \\

\end{tabular} & \begin{tabular}{@{}p{5em}@{}c@{}} 
to & \xmark \\
it & \xmark \\
been & \rmark \\
not & \xmark \\
won & \rmark \\
\end{tabular} & \begin{tabular}{@{}p{5em}@{}c@{}} 

won & \rmark \\
died & \rmark \\
left & \rmark \\
\textbf{gone} & \rmark \\
forgotten & \rmark \\
\end{tabular}  & I \ftoken{had} \otoken{gone}. & 
 \\



\hline
The clip is about a [MASK]. & \begin{tabular}{@{}p{5em}@{}c@{}} 
\textbf{minute} & \lmark \\
year & \lmark \\
second & \lmark \\
day & \lmark \\
week & \lmark \\
\end{tabular} & The clip is \ftoken{about} a \otoken{minute}.
  & \begin{tabular}{@{}p{5em}@{}c@{}} 

\textbf{minute} & \lmark \\
second & \lmark \\
third & \xmark \\
minutes & \xmark \\
moment & \lmark \\
  
\end{tabular} & \begin{tabular}{@{}p{5em}@{}c@{}}
\textbf{minute} & \lmark \\
second & \lmark \\
third & \xmark \\
minutes & \xmark \\
moment & \lmark \\
\end{tabular} & \begin{tabular}{@{}p{5em}@{}c@{}}

woman & \rmark \\
man & \rmark \\
\textbf{minute} & \lmark \\
day & \lmark \\
year & \lmark \\
\end{tabular} & \begin{tabular}{@{}p{5em}@{}c@{}}
woman & \rmark \\
girl & \rmark \\
man & \rmark \\
child & \rmark \\
boy & \rmark \\
\end{tabular}  & The clip is \ftoken{about} a \otoken{queen}. & 
 \\
\hline
The dinner is on the [MASK]. & \begin{tabular}{@{}p{5em}@{}c@{}} 
table & \lmark \\
rocks & \lmark \\
way & \xmark \\
beach & \lmark \\
menu & \xmark \\
\end{tabular} & The dinner is \ftoken{on} the \otoken{plate}.
& \begin{tabular}{@{}p{5em}@{}c@{}} 
table & \lmark \\
house & \xmark \\
menu & \xmark \\
line & \xmark \\
way & \xmark \\
\end{tabular} & \begin{tabular}{@{}p{5em}@{}c@{}} 
same & \xmark \\
usual & \xmark \\
winner & \xmark \\
evening & \xmark \\
weekend & \rmark \\
\end{tabular} & \begin{tabular}{@{}p{5em}@{}c@{}} 
evening & \xmark \\
weekend & \rmark \\
Sunday & \rmark \\
\textbf{Wednesday} & \rmark \\
village & \xmark \\
\end{tabular} & \begin{tabular}{@{}p{5em}@{}c@{}} 
\textbf{Wednesday} & \rmark \\
Sunday & \rmark \\
weekend & \rmark \\
same & \xmark \\
Saturday & \rmark \\
\end{tabular}  & The dinner is \ftoken{on} \otoken{Wednesday}. &
   \\

\hline

\end{tabular}
}
\caption{Example top-5 predictions with interpolated MaPP versus Vanilla BERT. \lmark{} indicates that the prediction has been coded as consistent with the Query~1 sense, \rmark{} for the Query~2 sense, and \xmark{} for neither. Predictions that result in an ungrammatical sentence are also coded as \xmark{}. The expectation is that values of $\alpha$ closer to 0 will be more reflective of the Query~1 sense, while values closer to 1 will be more reflective of the Query~2 sense. (Note that 0.4 and 0.8 are not evenly spaced between 0 and 1.) Original words from the queries are bolded.}
\label{tab:interpolation_examples}
\end{table*}

\paragraph{Results.} The fraction of predicted words consistent with the original sense decreases gradually as the amount of perturbation $\epsilon$ increases (\cref{plot_bar_epsilon_top1} and numeric results in \cref{app:more_analysis_results}). 
This matches our hypothesis that there is regularity in the BERT space, and that it is carved into regions which correspond to distinct senses. Outside of these regions (where $\epsilon$ is large), we occasionally encounter \emph{sense voids}---regions where there is no intelligible sense compatible with the context.
For example, with the query ``The event is \ftoken{in} \otoken{Canada}.'', 
we see small values of $\epsilon$ producing names of countries, but $\epsilon\geq 0.6$ producing adjectives (``annual'', ``amateur'', ``contested'', ``open'', ``free'') which are ungrammatical and nonsensical in context.


\subsection{Interpolation}

\begin{figure}
    \centering
    \includegraphics[width=0.45\textwidth]{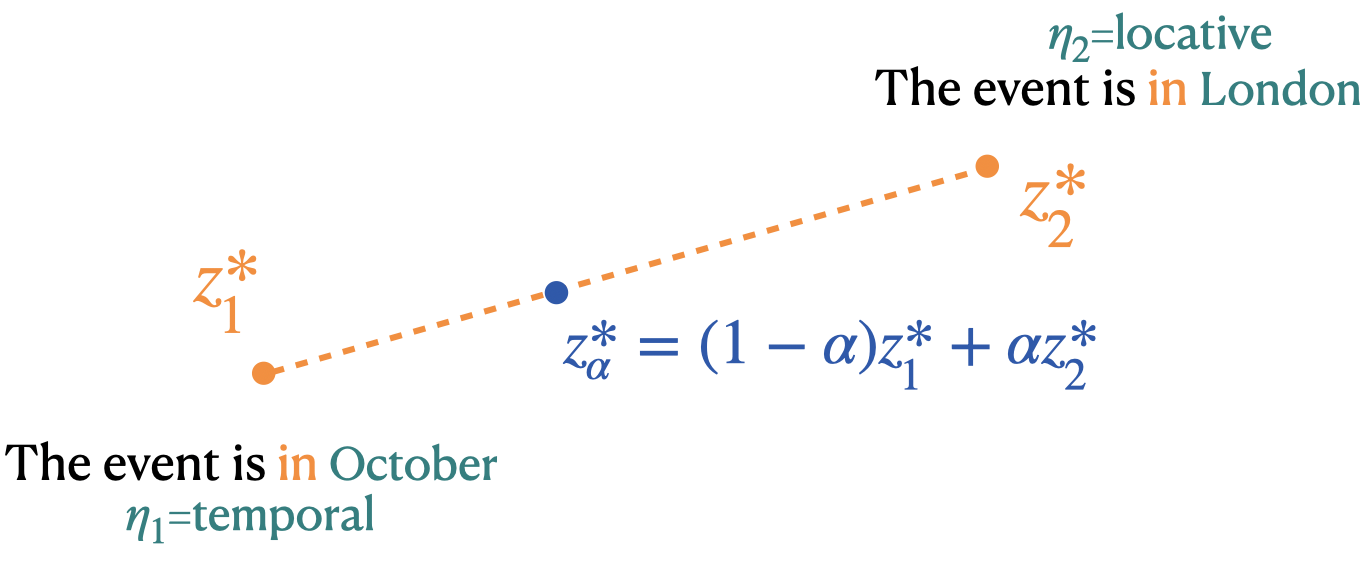}
    \caption{Illustration of the interpolation process. \vspace{-.5cm}}
    \label{fig:interpolation}
\end{figure}

Next, we take two pseudowords representing distinct senses of an ambiguous word in minimal pair sentences, 
and traverse the space between them
to determine what the boundary between sense regions looks like.
Given two pseudowords $\bz_1$ and $\bz_2$, we simply interpolate their vectors: 
${\bz^{*}_{\alpha}} = (1-\alpha) {\bz^{*}_1} + \alpha {\bz^{*}_2}$, 
where $0 \leq \alpha \leq 1$ controls how much weight to put on one pseudoword or the other.
This is depicted in \cref{fig:interpolation}.

\paragraph{Data.} In this experiment we use the \textbf{Minimal Pairs Portion} of our dataset. These $40$ pairs of sentences differ only in the cue to give contrasting senses of the focus word.\footnote{In some cases, the two elements of a minimal pair differ syntactically as well. Viz.: 
auxiliary vs.~main verb (``I \ftoken{had} \otoken{gone}/\otoken{cake}'');
verb-particle construction vs.~verb+PP (``run \ftoken{over} the \otoken{cat}/\otoken{bridge}''); and PP vs.~approximation modifier (``\ftoken{about} a \otoken{horse}/\otoken{minute}'').} $7$ different ambiguous words and $16$ distinct senses appear in this portion of the dataset, with $5$ sentences for each distinct sense. 
Several examples appear in \cref{table min pairs}.

\paragraph{Evaluation.} 
For each sentence in a minimal pair, we infer $\bz$. 
Then for $\alpha \in \{0,0.1,0.15,0.2,\dots,1\}$, we compute $\bz^{*}_{\alpha}$, use it for masked prediction, and judge whether each of the top 5 predictions corresponds to the sense in the first sentence, the second sentence, or neither.

\begin{figure}[t] 
\centering

\includegraphics[clip,trim=0 0 0 70,width=8cm]{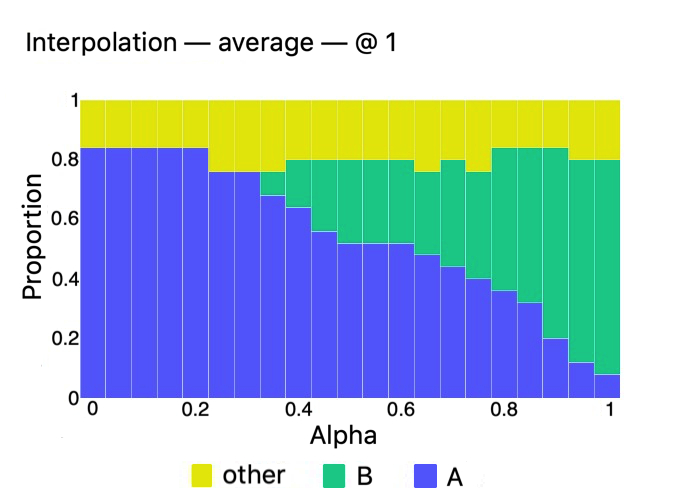}
\caption{Interpolation results for minimal pair data as a function of interpolation parameter $\alpha$: average proportion of top-1 predictions consistent with sense A, which predominates at $\alpha=0$; sense B, which predominates at $\alpha=1$; or neither.\vspace{-0.7cm}}
\label{plot_bar_alpha_top1}
\end{figure}

\paragraph{Results.} 
\Cref{plot_bar_alpha_top1} shows the overall proportion of predictions for each sense as $\alpha$ progresses from $0$ to $1$. We see a gradual trend from one sense to the other. This matches our hypothesis that there is regularity in the BERT-space: traveling on a line between two senses in the input spaces decodes to two distinct regions in the BERT-space.

For some individual examples, there is a sharp boundary at some $\alpha$; for others there is an intermediate region where the predictions mix the two senses or are unrelated to both (see \cref{app:more_analysis_results}).

The behavior with static embeddings (``Vanilla BERT'') can serve as a control for interpreting the effect of the interpolated pseudoword, as shown for several examples in \cref{tab:interpolation_examples}.
In many cases Vanilla BERT prefers one of the two senses by default, 
but with $\alpha$ sufficiently close to the other sense's pseudoword, 
the behavior changes.
In general, the transition from one sense to the other is readily apparent from the predictions.
Exceptions where one of the expected senses is inadequately represented include ``I \ftoken{had} \otoken{cake}.''\ (no foods are predicted in the top 5) and ``The dinner is \ftoken{on} the \otoken{plate}.''\ (not as many food-oriented locations were predicted as expected).

\subsection{Generalization Experiments}

In this experiment we examine whether the pseudoword is specialized for a particular sense of the focus word only in a context-specific fashion, 
or whether the pseudoword is a valid representation of the sense in new contexts where the focus word may appear.
To this end we take a pseudoword from one context and ``transplant'' it into a new context. We are particularly interested in transplantations where the ambiguous word has a similar meaning but is expected to yield a new distribution of masked predictions, due to the influence of the new context. For example:

\ex.\a. \label{for_purpose1} $s$: The book is \ftoken{for} \otoken{reading}.
    \b. \label{for_purpose2} $s'$: The cup is \ftoken{for} [MASK].
    \z.

Both \cref{for_purpose1} and \cref{for_purpose2} exemplify the `purpose of item' sense of \textbf{for}, for different kinds of items that have different kinds of ordinary purposes. If the pseudoword inferred from the original sentence appropriately generalizes the meaning of \textbf{for}, then transplanting it into $s'$ should yield a word like ``drinking''. However, in most cases the prediction either does not change (here, for example, we still get [MASK] = ``reading''), or we get an incoherent prediction for the masked token.

We hypothesize that this is because the pseudoword overfit to the original context---that is, it is incapable of representing the desired sense in new contexts (especially if the meaning of the new context crucially affects what should be predicted in the masked slot).

We hypothesize that it is necessary to take multiple contexts into account in order to produce a flexible-context sense-like vector. 
One possible strategy is to compute a sense-vector as a simple average of the individually-learned pseudoword.
We refer to this as \textbf{post hoc averaging}.

Another possible strategy is to train each pseudoword on multiple examples with distinct contexts: we replace \cref{loss} with an \textbf{aggregate loss} that averages over $n$ sentences containing the same focus token with the same sense $\eta$: 

\begin{equation} \label{loss_average}
\bz^{*}_{\eta} = \argmin_{\bz \in \mathbb{R}^{d}} \frac{1}{n} \sum_{i=1}^{n} {||BERT (\bz) - {\bx^{(i)}_t} ||}^2
\end{equation}

\noindent 

Note that both approaches inject supervision into the process of training the sense vector by specifying which examples correspond to the same sense.

\paragraph{Data.} In this experiment we use the \textbf{Generalizaton Portion} of the MaPP Dataset, which contains $138$ sentences with $5$ ambiguous words and $6$ distinct senses. For each sense, there are $23$ sentences (with $14$ sentences used as the training set to compute the averages, and $9$ as the test set).

\paragraph{Evaluation.} For each sentence we compute two $\bz^*$ pseudowords with the two kinds of averaging (post hoc and aggregate loss),  
and compare their effectiveness at adapting the expected sense for the new context.

In total we evaluate $270$ predictions.

\begin{table}[t] 
\centering
\small
\begin{tabular}{lrr}
\toprule
\textbf{generalization type} & \textbf{@1} & \textbf{@5} \\
$N$ (total \# of predictions) & 54  & 270 \\
\midrule 
Vanilla BERT baseline ($s'$ only) & 31.5 & 31.9 \\
MaPP: post hoc average & 11.1 & 14.8 \\
MaPP: aggregate loss (\cref{loss_average}) & 57.4 & 53.7 \\
\bottomrule
\end{tabular}
\caption{Generalization experiment. Comparison of $@ 1$ and $@ 5$ accuracy, over two generalization types; simple average of the pseudowords versus averaging in the loss function.\vspace{-.5cm}}
\label{table_generalization}
\end{table}

\paragraph{Results.}
\Cref{table_generalization} shows generalization accuracies for the two techniques as well as Vanilla BERT.
We see that the aggregate loss technique produces a correct prediction a majority of the time, while the static embedding is less accurate and post hoc averaging of pseudowords performs very poorly. 
These results support our intuition regarding the possibility to generate a representation for pseudowords that generalizes over different usages of the word. However, an ideal representation of a pseudoword would be one that could serve as a sense-disambiguated embedding of the focus word. This might not be completely achievable, but the representation might be improved in this direction by learning it over a larger more diverse dataset.

\section{Discussion}

\paragraph{What is a pseudoword?}
The optimization problem defined in \cref{loss} results in a pseudoword $\bz^* \in \mathbb{R}^{d}$. We use the pseudowords as input vectors to the model, although they are not constrained to the 30k vectors in BERT's vocabulary, but may be arbitrary vectors in $\mathbb{R}^{d}$. We discuss the validity of such an operation in \cref{app:further_discussion}. 

In practice, we find that many pseudowords behave as sense-disambiguated input vectors.
While our goal is not to explore the pseudoword-space for its own sake---pseudowords are a tool to shed light on the geometry and behavior of the BERT-space---our experiments with pseudowords and artificially perturbed pseudowords reveal that the pseudoword-space  
contains regions that are semantically coherent as inputs to BERT.


\paragraph{Prospects for the MaPP technique.} 
Our dataset is manually curated to control for specific linguistic phenomena. 
 
We expect that pseudoword may be less semantically targeted if learned with larger contexts that create more opportunities for confounds. 
We note also that senses are not necessarily discrete \citep{erk-mccarthy-2009-graded}, and it would be worthwhile to explore how graded semantic distinctions are represented, as well as underspecified meanings. We are also interested in exploring how BERT represents tokens in sentences that permit multiple plausible interpretations. 
The MaPP technique can be applied to investigate the properties of other CR models as well, as it requires only that the model be a differentiable function from input token embeddings to contextualized embeddings.


\section{Conclusion}

We have presented a novel methodology and a dataset for investigating the geometry of the BERT-space, using a traversal technique which allows for a continuation of the input space. 
We showed that there is substantial regularity in the BERT-space, 
with regions that correspond to distinct senses. Moreover, we found evidence for ``voids'' in the space---regions that do not correspond to any intelligible sense. 
Our technique gives rise to various types of analysis, creating avenues for future work. Immediate directions that we plan to pursue are (a)~examining sense representation in longer, naturally occurring sentences, and (b)~extending our analysis to a multilingual setting.

\finalversion{\section*{Acknowledgments}}

This work was supported by the Israel Science Foundation (grant no.~929/17).
Taelin Karidi was partially supported by a fellowship from the Hebrew University Center for Interdisciplinary Data Science Research.
Vivek Srikumar and Yichu Zhou were partly supported by NSF grants \#1801446 (SATC) and \#1822877 (Cyberlearning) and an award from Verisk Inc.
We appreciate feedback from reviewers, Sean Trott, and members of the NERT lab at Georgetown.




\bibliography{anthology,coersionBERT}
\bibliographystyle{acl_natbib}
\FloatBarrier

\appendix

\section{Appendix}

\subsection{The MaPP Dataset}

We present here the ambiguous words together with their senses, as used in the MaPP Dataset (\cref{app:mapp_dataset_table}). 

\begin{table}[ht]
\centering
\small
\begin{tabular}{>{\bf\ftoken\bgroup}r<{\egroup}@{~~~}p{15em}}
\toprule

\multicolumn{2}{c}{\textbf{Basic portion}} \\
	   about & topic, duration \\
       for  & duration, recipient, purpose \\
       had & auxiliary past participle, social event, food, medical condition \\
       in & locative, temporal \\
       on & locative, temporal \\
       run & manage, motion \\
       started & electronic device, information source \\
       with & instrument, feeling, accompanier \\

\midrule

\multicolumn{2}{c}{\textbf{Minimal pairs portion}} \\ 
	   about & topic, duration \\
       for  & recipient, purpose \\
       had & auxiliary past participle, social event, food, medical condition \\
       in & locative, temporal \\
       on & locative, temporal \\
       run & manage, motion \\
       started & electronic device, information source \\
       with & instrument, feeling \\

\midrule
\multicolumn{2}{c}{\textbf{Generalization portion}} \\ 
	   about & topic \\
       for  & purpose  \\
       had & auxiliary past participle, food \\
       with & feeling, accompanier \\

\bottomrule

\end{tabular}
\caption{Ambiguous words and senses for the different portions of the MaPP Dataset.}
\label{app:mapp_dataset_table}
\end{table}

\FloatBarrier


\subsection{Relational Words As a Test Case} \label{app:relational}

We chose to focus our analysis on relational words ~\cref{sec:dataset}. 
Understanding what is encoded in representations of these words can shed light on some of the open questions regarding the semantic and syntactic knowledge that is encoded in CRs.
From prior works on classification of relational words with CRs~\cite[e.g.,][]{liu-etal-2019-linguistic}, we know that these differences (in the sense and the form of relational words) are indeed encoded in them. Indeed, in some settings, it is possible to separate groups of them via simple classifiers. However, this is a weak notion of the knowledge that is encoded in the representation.
Other work that focused on probing for function words comprehension~\cite{kim-etal-2019-probing} explored whether qualitatively different objectives lead to demonstrably different sentence representations.
To understand to what extent (and how) the form of a word versus its sense is encoded in its contextual representation, we have conducted the experiments that we describe in Section $5$.

\subsection{Our method vs. Other Probing Methods}

Our work addresses two basic shortcomings of most probing methods. First, they strongly rely on the probing dataset used to train and evaluate a classifier. Changing the distribution of examples can shift the results of the probing experiment~\cite[e.g.,][]{slobodkin-etal-2021-mediators}. Second, probing methods give an aggregated picture at the population level, and cannot provide insight at the level of individual examples. 

Our method does not train a classifier, and can provide information at the instance level; it therefore does not rely on aggregation  to yield a meaningful conclusion. Rather, it is designed to allow for an interpretable navigation of the BERT space. While our method does allow reporting trends at the population level by aggregation, results can be traced back to the instance level.

\subsection{Further Discussion} \label{app:further_discussion}
\paragraph{Transfer learning using BERT.} 

Although BERT is built as a masked language model, it is often being used as a tool for transfer learning; its produced representations are treated as vectors in a continuous space and are being used with great success for various tasks such as POS tagging, NLI, multilingual alignments, prediction of brain activity patterns, and more \cite{schuster-etal-2019-cross, gauthier-levy-2019-linking,rogers-etal-2020-primer}. 

Our method uses pseudowords as input vectors to the model. However, the vectors that are given as an input to BERT are always one of $30k$ vectors in BERT's base vocabulary, where any other vector is considered ``out of vocabulary''. BERT was never meant to receive any vector in $\mathbb{R}^{d}$ as it is defined over a discrete set, yet we are breaking the "discrete-contract" and asking what is the behaviour of the model given the pseudowords and perturbations of them. To the best of our knowledge this is a novel approach to the exploration of BERT. If BERT was treated only as a masked language model then one could claim that there is a certain set of rules that is needed to be followed in order to infer meaningful conclusions from its outputs. In our approach however, we choose adopt a different view -- we think about BERT as a sentence encoder; a function from a sequence of strings to sequence of vectors. We claim that in adopting this approach there is no need to constrain the model to a discrete space. Moreover, this "contract" has in fact already been violated, as contextualized representations are often being used for other tasks other than masked language modeling, and therefore the use of pseudowords as inputs is nothing short of a natural continuation of this idea.
\quash{Moreover, as contextualized representations are often being used for other tasks than masked language modeling, this "contract" is in fact violated, and therefore the use of pseudowords as inputs is nothing short of a natural continuation of this concept.}  


\subsection*{More Analysis Results} \label{app:more_analysis_results}

\quash{

\begin{table}[!htb] 
\centering
\small
\resizebox{\columnwidth}{!}{
\begin{tabular}{@{}lrrrrrrrrrr@{}}
\toprule
& \multicolumn{2}{c}{\textbf{$0 \leq \alpha \leq 0.2$}} & \multicolumn{2}{c}{\textbf{$0.25 \leq \alpha \leq 0.4$}} & \multicolumn{2}{c}{\textbf{$0.45 \leq \alpha \leq 0.6$}} &
\multicolumn{2}{c}{\textbf{$0.65 \leq \alpha \leq 0.8$}} &
\multicolumn{2}{c@{}}{\textbf{$0.85 \leq \alpha \leq 1$}} \\
 \cmidrule(lr){2-3}\cmidrule(lr){4-5}\cmidrule(lr){6-7}\cmidrule(lr){8-9}\cmidrule(l){10-11}
 & \textbf{@1} & \textbf{@5} & \textbf{@1} & \textbf{@5} & \textbf{@1} & \textbf{@5} & \textbf{@1} & \textbf{@5} & \textbf{@1} & \textbf{@5} \\
$N$ & 200 & 1k & 200 & 1k & 200 & 1k & 200 & 1k & 240 & 1.2k \\
\midrule
Sense A & 84.0 & 81.0 & 76.0 & 74.8 & 56.0 & 48.8 & 46.0 & 32.2 & 21.6 & 18.1 \\
Sense B & 0.0 & 0.0 & 2.0 & 5.4 & 24.0 & 29.2 & 32.0 & 41.8 & 60.8 & 57.4 \\
Other & 16.0 & 19.0 & 22.0 & 19.8 & 20.0 & 22.0 & 22.0 & 25.6 & 17.6 & 23.5 \\

\bottomrule
\end{tabular}
}
\caption{Interpolation experiment. Comparison of $@1$ and $@5$ accuracy for sense A (start), sense B (end) and OTHER, over intervals of $\alpha$.}

\label{table_interpolation}
\end{table}
}

\FloatBarrier

\begin{figure}[!h]
  \includegraphics[trim=0 50 0 25,clip,width=\columnwidth]{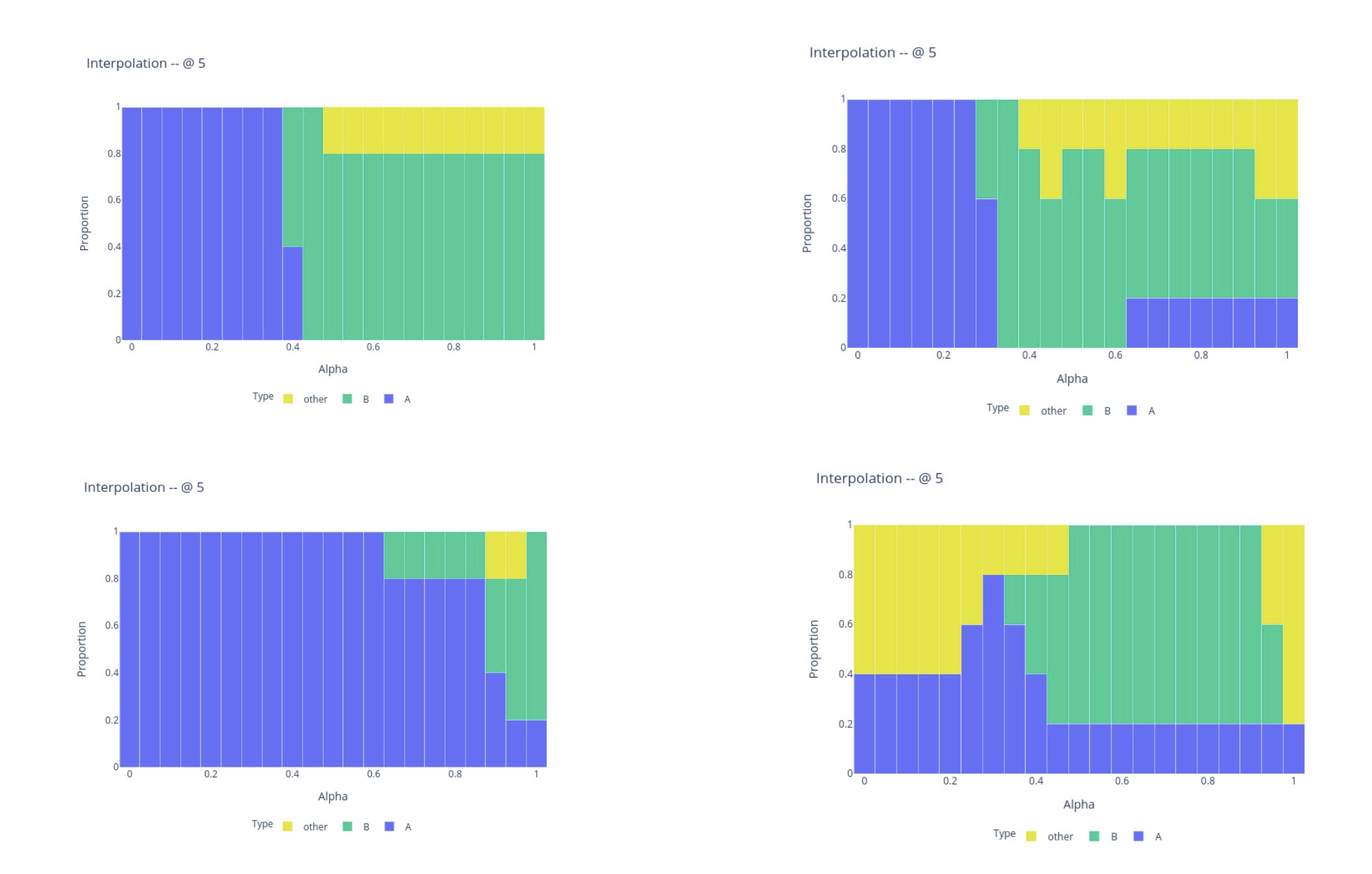}
  \caption{Interpolation results for individual minimal pairs: top-5 sense matches over the range of $\alpha$ values, plotted in the same style as Figure $5$. These cases are quite rare and provide examples for departures from the main trend.}
  \label{app:fig_inter1}
\end{figure}

\FloatBarrier


\begin{table}[!h] 
\centering
\small
\resizebox{\columnwidth}{!}{
\begin{tabular}{@{}lrrrrrr@{}}
\toprule
& \multicolumn{2}{c}{\textbf{$0 \leq \epsilon \leq 0.4$}} & \multicolumn{2}{c}{\textbf{$0.6 \leq \epsilon \leq 1$}} & \multicolumn{2}{c@{}}{\textbf{$1.2 \leq \epsilon \leq 1.8$}} \\
 \cmidrule(lr){2-3}\cmidrule(lr){4-5}\cmidrule(l){6-7}
 & \textbf{@1} & \textbf{@5} & \textbf{@1} & \textbf{@5} & \textbf{@1} & \textbf{@5} \\
$N$ (total \# of predictions) & 2.82k & 4.7k & 2.82k & 4.7k & 3.76k & 18.8k \\
\midrule
Same sense ($z^*$) & 74.9 & 63.9 & 36.3 & 32.5 & 14.1 & 15.7 \\
   
\bottomrule
\end{tabular}
}
\caption{$\epsilon$-perturbation experiment. Comparison of $@1$ and $@5$ accuracy over intervals of $\epsilon$. 
}
\label{table_epsilon_perturbation}
\end{table}

\FloatBarrier

\begin{table*}[!t]
\centering
\footnotesize
\resizebox{\linewidth}{!}{
\begin{tabular}{|c|c|c|c|c|c|c|}
	\hline

\multicolumn{1}{|c|}{\multirow{2}{*}{\textbf{Mask}}} & \multicolumn{1}{c|}{\multirow{2}{*}{\textbf{Vanilla BERT}}} & \multirow{2}{*}{\textbf{Query}} & \multicolumn{4}{|c|}{\textbf{MaPP}} \\ \cline{4-7}
	
	\multicolumn{1}{|c|}{} & \multicolumn{1}{|c|}{} & \multicolumn{1}{|c|}{} & \textbf{$\epsilon = 0$} & \textbf{$\epsilon = 0.6$} & \textbf{$\epsilon = 1.2$} &  \textbf{$\epsilon = 1.8$} 	\\ \hline

The event is in [MASK]. &
\begin{tabular}{@{}ll@{}} 
progress & \xmark \\
June & \xmark \\
July & \xmark \\
April & \xmark \\
September & \xmark \\
\end{tabular} &
The event is {\color{orange} in} {\color{teal} London}. &
\begin{tabular}{@{}ll@{}} 
London & \cmark \\
Dublin & \cmark \\
Edinburgh & \cmark \\
Paris & \cmark \\
Sydney & \cmark \\
\end{tabular} & \begin{tabular}{@{}ll@{}} 
London & \cmark \\
Dublin & \cmark \\
Toronto & \cmark \\
Melbourne & \cmark \\
Sydney & \cmark \\
\end{tabular} & \begin{tabular}{@{}ll@{}} 
{\#}{\#}e & \xmark \\
{\#}{\#}a & \xmark \\
{\#}{\#}actic & \xmark \\
{\#}{\#}atic & \xmark \\
free & \xmark \\
\end{tabular} & \begin{tabular}{@{}ll@{}} 
{\#}{\#}e & \xmark \\
{\#}{\#}ental & \xmark \\
{\#}{\#}ated & \xmark \\
{\#}{\#}an & \xmark \\
{\#}{\#}anche & \xmark \\
\end{tabular} \\

\hline
It lasted for [MASK]. &
\begin{tabular}{@{}ll@{}} 
hours & \cmark \\
days & \cmark \\
awhile & \cmark \\
months & \cmark \\
weeks & \cmark \\
\end{tabular} & 
It lasted {\color{orange} for} {\color{teal} seconds}. &
\begin{tabular}{@{}ll@{}} 
hours & \cmark \\
minutes & \cmark \\
seconds & \cmark \\
awhile & \cmark \\
forever & \xmark \\
\end{tabular} & \begin{tabular}{@{}ll@{}} 
minutes & \cmark \\
seconds & \cmark \\
hours & \cmark \\
days & \cmark \\
years & \cmark \\
\end{tabular} & \begin{tabular}{@{}ll@{}} 
{\#}{\#}y & \xmark \\
minutes & \cmark \\
long & \xmark \\
{\#}{\#}ily & \xmark \\
seconds & \cmark \\
\end{tabular} & \begin{tabular}{@{}ll@{}} 
forever & \xmark \\
hours & \cmark \\
awhile & \cmark \\
minutes & \cmark \\
longer & \cmark \\
\end{tabular} \\

\hline
The book is for [MASK]. &
\begin{tabular}{@{}ll@{}} 
children & \xmark \\
women & \xmark \\
adults & \xmark \\
sale & \cmark \\
boys & \xmark \\
\end{tabular} &
The book is {\color{orange} for} {\color{teal} learning}. &
\begin{tabular}{@{}ll@{}} 
reading & \cmark \\
children & \xmark \\
learning & \cmark \\
education & \cmark \\
use & \cmark \\
\end{tabular} & \begin{tabular}{@{}ll@{}} 
children & \xmark \\
reading & \cmark \\
sale & \cmark \\
women & \xmark \\
adults & \xmark \\
\end{tabular} & \begin{tabular}{@{}ll@{}} 
{\#}{\#}o & \xmark \\
{\#}{\#}olate & \xmark \\
{\#}{\#}aged & \xmark \\
{\#}{\#}olic & \xmark \\
free & \xmark \\
\end{tabular} & \begin{tabular}{@{}ll@{}} 
bilingual & \xmark \\
free & \xmark \\
incomplete & \xmark \\
lost & \xmark \\
anonymous & \xmark \\
\end{tabular} \\

\hline
I started the [MASK]. &
\begin{tabular}{@{}ll@{}} 
engine & \cmark \\
car & \cmark \\
motor & \cmark \\
truck & \cmark \\
ignition & \cmark \\
\end{tabular} &
I {\color{orange} started} the {\color{teal} bus}. &
\begin{tabular}{@{}ll@{}} 
engine & \cmark \\
car & \cmark \\
bike & \cmark \\
truck & \cmark \\
motor & \cmark \\
\end{tabular} & \begin{tabular}{@{}ll@{}} 
car & \cmark \\
truck & \cmark \\
engine & \cmark \\
bus & \cmark \\
bike & \cmark \\
\end{tabular} & \begin{tabular}{@{}ll@{}} 
bird & \xmark \\
phone & \xmark \\
same & \xmark \\
birds & \xmark \\
car & \cmark \\
\end{tabular} & \begin{tabular}{@{}ll@{}} 
same & \xmark \\
bird & \xmark \\
hell & \xmark \\
other & \xmark \\
number & \xmark \\
\end{tabular} \\

\hline
I had [MASK]. &
\begin{tabular}{@{}ll@{}} 
to & \xmark \\
it & \xmark \\
nothing & \xmark \\
him & \xmark \\
her & \xmark \\
\end{tabular} &
I {\color{orange} had} {\color{teal} slept}. &
\begin{tabular}{@{}ll@{}} 
slept & \cmark \\
forgotten & \cmark \\
not & \xmark \\
been & \cmark \\
died & \cmark \\
\end{tabular} & \begin{tabular}{@{}ll@{}} 
forgotten & \cmark \\
to & \xmark \\
been & \cmark \\
slept & \cmark \\
{\#}{\#}ed & \xmark \\
\end{tabular} & \begin{tabular}{@{}ll@{}} 
asked & \cmark \\
{\#}{\#}o & \xmark \\
nodded & \cmark \\
{\#}{\#}a & \xmark \\
swallowed & \cmark \\
\end{tabular} & \begin{tabular}{@{}ll@{}} 
asked & \cmark \\
said & \cmark \\
nodded & \cmark \\
ask & \xmark \\
smiled & \cmark \\
\end{tabular} \\

\hline
I had a [MASK]. &
\begin{tabular}{@{}ll@{}} 
plan & \xmark \\
point & \xmark \\
headache & \xmark \\
feeling & \xmark \\
choice & \xmark \\
\end{tabular} &
I {\color{orange} had} a {\color{teal} reception}. &
\begin{tabular}{@{}ll@{}} 
meeting & \cmark \\
surprise & \xmark \\
party & \cmark \\
call & \xmark \\
date & \cmark \\
\end{tabular} & \begin{tabular}{@{}ll@{}} 
lot & \xmark \\
smile & \xmark \\
look & \xmark \\
feeling & \xmark \\
friend & \xmark \\
\end{tabular} & \begin{tabular}{@{}ll@{}} 
lot & \xmark \\
little & \xmark \\
thought & \xmark \\
feeling & \xmark \\
smile & \xmark \\
\end{tabular} & \begin{tabular}{@{}ll@{}} 
little & \xmark \\
nod & \xmark \\
thought & \xmark \\
sigh & \xmark \\
guess & \xmark \\
\end{tabular} \\

\hline
The clip is about a [MASK]. &
\begin{tabular}{@{}ll@{}} 
minute & \xmark \\
year & \xmark \\
second & \xmark \\
day & \xmark \\
week & \xmark \\
\end{tabular} &
The clip is {\color{orange} about} a {\color{teal} queen}. &
\begin{tabular}{@{}ll@{}} 
woman & \cmark \\
girl & \cmark \\
man & \cmark \\
child & \cmark \\
boy & \cmark \\
\end{tabular} & \begin{tabular}{@{}ll@{}} 
woman & \cmark \\
girl & \cmark \\
man & \cmark \\
mountain & \cmark \\
city & \cmark \\
\end{tabular} & \begin{tabular}{@{}ll@{}} 
shot & \xmark \\
cartoon & \cmark \\
song & \cmark \\
picture & \cmark \\
photograph & \cmark \\
\end{tabular} & \begin{tabular}{@{}ll@{}} 
video & \cmark \\
photograph & \cmark \\
cartoon & \cmark \\
documentary & \cmark \\
picture & \cmark \\
\end{tabular} \\

\hline
The clip is about a [MASK]. &
\begin{tabular}{@{}ll@{}} 
minute & \cmark \\
year & \cmark \\
second & \cmark \\
day & \cmark \\
week & \cmark \\
\end{tabular} &
The clip is {\color{orange} about} a {\color{teal} minute}. &
\begin{tabular}{@{}ll@{}} 
minute & \cmark \\
second & \cmark \\
third & \xmark \\
minutes & \xmark \\
moment & \cmark \\
\end{tabular} & \begin{tabular}{@{}ll@{}} 
minute & \cmark \\
year & \cmark \\
second & \cmark \\
day & \cmark \\
week & \cmark \\
\end{tabular} & \begin{tabular}{@{}ll@{}} 
documentary & \xmark \\
ballad & \xmark \\
video & \xmark \\
photograph & \xmark \\
film & \xmark \\
\end{tabular} & \begin{tabular}{@{}ll@{}} 
documentary & \xmark \\
video & \xmark \\
photograph & \xmark \\
diary & \xmark \\
ballad & \xmark \\
\end{tabular} \\

\hline
The dinner is on the [MASK]. &
\begin{tabular}{@{}ll@{}} 
table & \cmark \\
rocks & \cmark \\
way & \xmark \\
beach & \cmark \\
menu & \xmark \\
\end{tabular} &
The dinner is {\color{orange} on} the {\color{teal} counter}. &
\begin{tabular}{@{}ll@{}} 
table & \cmark \\
counter & \cmark \\
floor & \cmark \\
nightstand & \cmark \\
kitchen & \xmark \\
\end{tabular} & \begin{tabular}{@{}ll@{}} 
table & \cmark \\
floor & \cmark \\
kitchen & \xmark \\
counter & \cmark \\
menu & \xmark \\
\end{tabular} & \begin{tabular}{@{}ll@{}} 
same & \xmark \\
best & \xmark \\
opposite & \xmark \\
first & \xmark \\
truth & \xmark \\
\end{tabular} & \begin{tabular}{@{}ll@{}} 
same & \xmark \\
following & \xmark \\
first & \xmark \\
winner & \xmark \\
result & \xmark \\
\end{tabular} \\

\hline
The dinner is on [MASK]. &
\begin{tabular}{@{}ll@{}} 
fire & \xmark \\
offer & \xmark \\
sale & \xmark \\
Friday & \cmark \\
hold & \xmark \\
\end{tabular} &
The dinner is {\color{orange} on} {\color{teal} Monday}. &
\begin{tabular}{@{}ll@{}} 
Sunday & \cmark \\
Saturday & \cmark \\
Thursday & \cmark \\
Tuesday & \cmark \\
Friday & \cmark \\
\end{tabular} & \begin{tabular}{@{}ll@{}} 
free & \xmark \\
open & \xmark \\
served & \xmark \\
prepared & \xmark \\
closed & \xmark \\
\end{tabular} & \begin{tabular}{@{}ll@{}} 
free & \xmark \\
open & \xmark \\
served & \xmark \\
closed & \xmark \\
private & \xmark \\
\end{tabular} & \begin{tabular}{@{}ll@{}} 
free & \xmark \\
private & \xmark \\
open & \xmark \\
served & \xmark \\
delicious & \xmark \\
\end{tabular} \\

\hline

\end{tabular}
}
\caption{ Examples top-5 prediction with $\epsilon$-Perturbubed MaPP versus vanilla BERT. $\cmark$ indicated that the prediction has been coded as consistent with Query sense, $\xmark$ for wrong prediction. The expectation is that values of $\epsilon$ closer to 0 will be more reflective of the Query sense, while as $\epsilon$ increases will be more incoherent or will not fit the Query sense}
\end{table*}

\FloatBarrier


\end{document}